\documentclass{article}
\usepackage{main}

\usepackage{times}
\usepackage{soul}
\usepackage{url}
\usepackage[hidelinks]{hyperref}
\usepackage[utf8]{inputenc}
\usepackage[small]{caption}
\usepackage{graphicx}
\usepackage{amsmath}
\usepackage{amsthm}
\usepackage{booktabs}
\usepackage{algorithm}
\usepackage{algorithmic}
\usepackage[switch]{lineno}

\usepackage{multirow}
\usepackage{amsfonts}
\DeclareMathOperator*{\argmin}{arg\,min}
\newcommand{\bfD}[0]{\mathbf{D}}
\newcommand{\bfX}[0]{\mathbf{X}}
\newcommand{\bfV}[0]{\mathbf{V}}
\newcommand{\calC}[0]{\mathcal{C}}
\newcommand{\calG}[0]{\mathcal{G}}
\newcommand{\calL}[0]{\mathcal{L}}
\newcommand{\calH}[0]{\mathcal{H}}
\newcommand{\calN}[0]{\mathcal{N}}
\newcommand{\doubleE}[0]{\mathbb{E}}
\newcommand{\etal}[1]{#1 \emph{et al.}}

\title{Generative Modeling for Robust Deep Reinforcement Learning\\on the Traveling Salesman Problem}

\author{
Michael Li$^1$\and
Eric Bae$^{1,*}$\and
Christopher Haberland$^{1,*}$\And
Natasha Jaques$^1$\\
\affiliations
$^1$University of Washington\\
$^*$Equal Contribution
\emails
\{ml10872, ebae3, haberc\}@uw.edu,
nj@cs.washington.edu
}

\begin{document}

\maketitle

\begin{abstract}
    The Traveling Salesman Problem (TSP) is a classic NP-hard combinatorial optimization task with numerous practical applications. Classic heuristic solvers can attain near-optimal performance for small problem instances, but become computationally intractable for larger problems. Real-world logistics problems such as dynamically re-routing last-mile deliveries demand a solver with fast inference time, which has led researchers to investigate specialized neural network solvers. However, neural networks struggle to generalize beyond the synthetic data they were trained on. In particular, we show that there exist TSP distributions that are realistic in practice, which also consistently lead to poor worst-case performance for existing neural approaches. To address this issue of distribution robustness, we present Combinatorial Optimization with Generative Sampling (COGS), where training data is sampled from a generative TSP model. We show that COGS provides better data coverage and interpolation in the space of TSP training distributions. We also present TSPLib50, a dataset of realistically distributed TSP samples, which tests real-world generalization ability without conflating this issue with instance size. We evaluate our method on various synthetic datasets as well as TSPLib50, and compare to state-of-the-art neural baselines. We demonstrate that COGS improves distribution robustness, with most performance gains coming from worst-case scenarios. \footnote{Our source code is available at \hyperlink{https://github.com/ML72/Generative-Model-TSP}{github.com/ML72/Generative-Model-TSP}.}
\end{abstract}

\section{Introduction}

\begin{figure*}[t]
    \centering
    \includegraphics[width=1.0\textwidth]{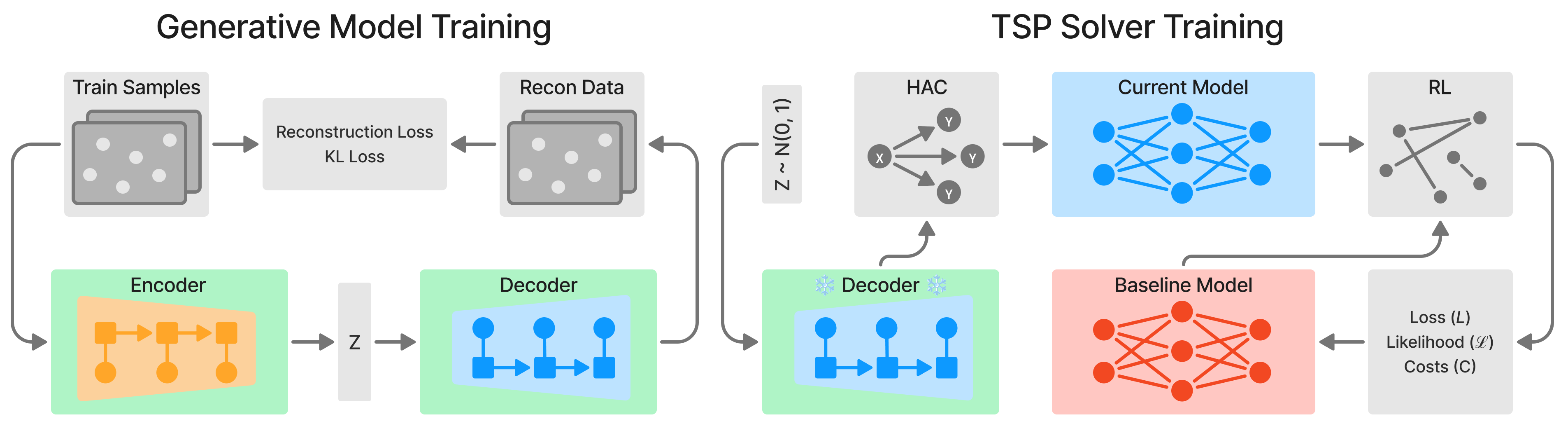}
    \caption{Architecture of our proposed Combinatorial Optimization with Generative Sampling (COGS) system. The generative model is first trained (left) and then frozen and used to generate training data for a deep reinforcement learning TSP solver (right).}
    \label{fig:architecture}
\end{figure*}

You are a well-known researcher who is invited to give talks at a number of famous universities around the world. The distance between universities directly correlates with the cost of flights between them. Your department offers to pay for your flights, but wishes to order the universities such that the total cost of your world tour is minimized. You suddenly realize that the problem at hand is the Traveling Salesman Problem (TSP), a well-known NP-hard combinatorial optimization problem (COP). From least-cost shipping and warehouse logistics to the efficient automated drilling of printed circuit boards, the TSP has an outsized impact on global trade, accounting for billions of dollars' worth of saved time, energy, human resources, and harmful emissions.

The TSP is NP-hard, which means there exists no efficient algorithm for finding exact solutions. While classic heuristic-based methods can provide near-optimal solutions for problems containing up to thousands of nodes, their runtime is prohibitive for situations requiring fast and dynamic decision-making. Neural combinatorial optimization (NCO) methods have sought to provide approximate solutions to the TSP at lower computational cost, and have demonstrated effectiveness \cite{miki_2018_applying,kool_2018_attention,shi_2022_nnsurvey,zaheer_2018_deepsets}. However, one general problem of NCO approaches in comparison to traditional heuristics is that they generalize poorly to unfamiliar distributions \cite{lisicki_2020_evaluating,fu_2021_generalization}. In this paper, we show that there exist realistic TSP distributions on which current neural approaches consistently perform highly suboptimally. In practice, such planning faults can be very expensive in terms of wasted time, money, and human resources. Lack of tight worst-case guarantees can also hinder the adoption of NCO in the real world.

A primary source of this bias is the data that current NCO models are trained on: namely, most current NCO approaches sample training data from uniform distributions. One possible solution is to train on randomly modified samples \cite{tobin_2017_domainrand}, but this is inefficient and wasteful in the high-dimensional space of possible TSP distributions. Curriculum-based learning methods, where agents are trained on progressively more difficult problem instances, have shown promise in combating this \cite{manchanda_2022_generalization,ouyang_2021_generalization}. For the TSP specifically, \etal{Zhang} \shortcite{zhang_2022_learning} proposed a hardness-adaptive curriculum (HAC), which uses gradient ascent to produce increasingly harder instances, and show that performance improves on harder distributions that are unknown to the model during training. However, since HAC relies on updating TSP instances sampled from a uniform distribution with a limited gradient step, the instances it trains on cannot diverge far the uniform distribution.

In this paper, we seek to improve the robustness of deep reinforcement learning TSP models to varying distributions. We present Combinatorial Optimization with Generative Sampling (COGS), an optimization approach which maintains the computational benefits of NCO methods while improving generalization on disparate but practical distributions. We construct a generative model for TSP distributions and then sample data from it to train our deep reinforcement learning TSP solver. The generative model's latent space properties lead to better coverage and interpolation across possible TSP distributions. Our method is depicted in Figure \ref{fig:architecture}. Our method is also relatively efficient during training and inference, requiring only a single GPU and no more than a few hours of training for a single model. This allows for significant cost reductions in practical applications, especially dynamic settings that require solving unknown configurations on the fly. Additionally, we propose the TSPLib50 dataset, bootstrapped from real-world distributions in the widely used TSPLib dataset \cite{reinhelt_2014_tsplib}, as a benchmark for model robustness on distributions of practical interest.

In summary, we make the following contributions:
\begin{itemize}
    \item We propose TSPLib50, a testing dataset of 10,000 instances sampled from realistic distributions.
    \item We propose a generative model for TSP distributions, which can interpolate between training samples, and infinitely generate new TSP problems that enable training a robust solver.
    \item We propose Combinatorial Optimization with Generative Sampling (COGS), an NCO training approach which results in improved robustness on distributions of practical interest.
\end{itemize}

\section{Background}

The Traveling Salesman Problem (TSP) is an NP-Complete combinatorial optimization problem, which requires finding the shortest tour through a set of locations \cite{papadimitriou_1977_euclideantsp}. More formally, given a set of locations $V = \{1, 2, \ldots, n\}$ and a $n\times n$ distance matrix $D$ where $D_{i,j}$ is a real number denoting the distance between locations $i$ and $j$, we seek to find the permutation of locations $\sigma^*$ that optimizes total tour length \cite{flood_1956_tsp}: 

\begin{equation}
    \sigma^* = \argmin_{\sigma} \left[D_{\sigma(n), \sigma(1)} + \sum_{i=1}^{n-1} D_{\sigma(i), \sigma(i+1)}\right]
    \label{eq:oracle_definition}
\end{equation}

The 2D-Euclidean TSP is a special case of the TSP where all locations are given a 2D position on the Euclidean plane, and all $D_{i,j}$ represent Euclidean distances between locations $i$ and $j$. Because $\sigma^*$ is translation-invariant and scale-invariant, 2D-Euclidean TSP problems often provide locations that are normalized to fit in the $[0,1]^2$ unit square.

Optimality gap (``gap" for short) is a measure of suboptimality relative to the optimal tour, and is the standard measure of TSP solver performance. Gap is typically expressed as a percentage, and lower gap is better. On datasets, gap is reported as an average over all instances. Formally, the gap of a model $M$ relative to an oracle $M^*$ on some dataset $\bfX$ is:

\begin{equation}
    \calG(\bfX, M) = \frac{\calC_{M} (\bfX) - \calC_{M^*} (\bfX)}{\calC_{M^*} (\bfX)}
    \label{eq:optimality_gap}
\end{equation}

The Lin-Kernighan algorithm \cite{lin_1973_lk} is regarded as a state-of-the-art near-optimal heuristic algorithm \cite{helsgaun_2000_effective,helsgaun_2017_extension}. Concorde is a widely used TSP exact solver \cite{applegate_2006_traveling}. In this paper, we use Concorde to compute optimality gaps.

\section{Related Work}

\subsection{Deep and Reinforcement Learning for TSP}

Neural combinatorial optimization (NCO), or the use of deep learning for combinatorial optimization, can be broadly grouped into three primary approaches: solutions utilizing 1) pointer networks, 2) graph neural networks, or 3) transformers \cite{liu_2023_good}. Pointer networks use an attention mechanism to select members of the input sequence as an output, and allows generalization to input and output lengths beyond training limits \cite{vinyals_2017_pointer,ma_2019_pointer}. Graph neural networks \cite{joshi_2019_efficient,min_2024_unsupervised} better allow dependencies between input elements to be considered \cite{zhou_2021_graph}. Transformer-based approaches build on top of encoder-decoder architectures using the attention mechanism \cite{vaswani_2017_attention}.

Reinforcement learning (RL) has seen successful applications in learning to solve the TSP \cite{mazyavkina_2021_reinforcement,deudon_2018_polgrad}. Deep RL methods often use a neural network to generate a tour, and then treat tour length as a negative reward, or ``cost". \etal{Kool} \shortcite{kool_2018_attention} propose a transformer-based solver trained with REINFORCE \cite{williams_1992_simple}, using a simple deterministic greedy rollout baseline. Their system attains near-optimal performance for TSP instances of up to 100 nodes when sampling test data from training distributions.

However, neural networks are known to often generalize poorly to distributions outside their training data, and existing NCO solvers are no exception. This makes them a risky solution for real-world deployments, in spite of their fast inference time. In this paper, we aim to improve the reliability and robustness of deep RL approaches.

\subsection{Curriculum Learning for TSP}

Curriculum methods are designed to improve robustness and sample efficiency by proposing tasks to learn from which are optimal for learning \cite{wang_2022_zone,azad_2023_unsupervisedcurriculum,dennis_2021_emergent}, and have been applied to real-world problems such as web navigation \cite{gur_2022_envgen}, and even other combinatorial optimization problems \cite{feng2020novel}. In the context of deep RL TSP solvers, \etal{Zhang} \shortcite{zhang_2022_learning} propose a hardness-adaptive curriculum (HAC), which mainly consists of two components: a hardness-adaptive generator that conducts gradient ascent on training instances, and a re-weighting procedure for batch gradients in favor of updates for harder instances. The hardness metric $\calH$ for a dataset $\bfX$ used by the generator is the gap in cost $\calC$ between the current model $M$ and a surrogate model $M'$ \cite{zhang_2022_learning}:

\begin{equation}
    \calH(\bfX, M) = \frac{\calC_{M} (\bfX) - \calC_{M'} (\bfX)}{\calC_{M'} (\bfX)}
    \label{eq:hardness_measurement}
\end{equation}

The hardness-adaptive generator, inspired by Langevin Dynamics, conducts gradient ascent on input samples $\bfX^{(t)}$ given a model $M$ \cite{zhang_2022_learning}:

\begin{equation}
    \bfX^{(t)'} = \bfX^{(t)} + \eta\nabla_{X^{(t)}} \calH(X^{(t)},M)
    \label{eq:hardness_adaptive}
\end{equation}

\etal{Zhang} \shortcite{zhang_2022_learning} also find that Gaussian Mixture distributions tend to be challenging for solvers trained on uniform distributions, and use such distributions as a benchmark for the effectiveness of HAC. Because HAC currently has the best results for improving performance on difficult distributions, we build off and compare to HAC in this paper.

\section{Preliminary Study}

\begin{figure}[t]
    \includegraphics[width=0.35\textwidth]{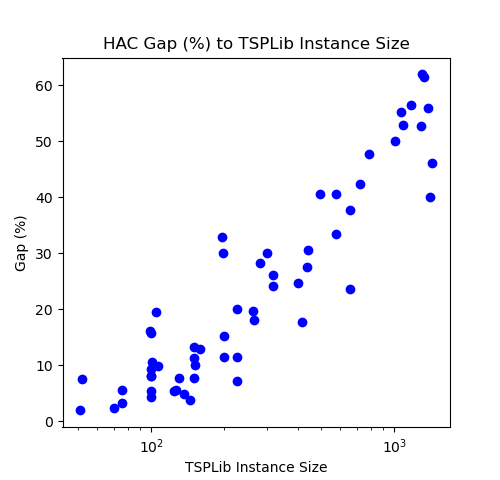}
    \centering
    \caption{There is a Pearson correlation of $0.907$ between HAC gap and TSPLib instance size. All 2D-Euclidean TSPLib instances with 1400 or fewer nodes are included.}
    \label{fig:prelim_hac_lengthgen}
\end{figure}

\begin{figure*}[t]
    \includegraphics[width=1.0\textwidth]{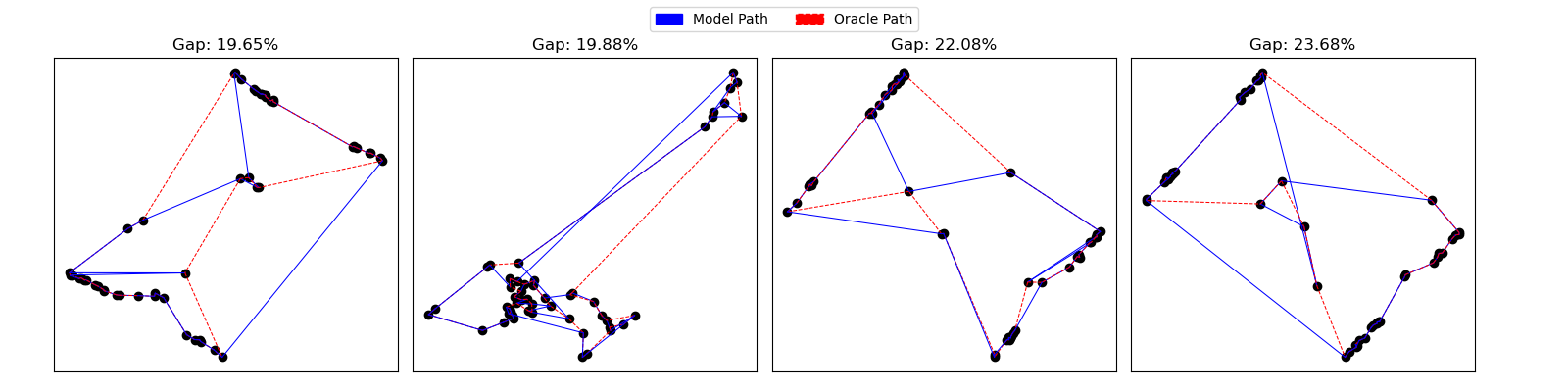}
    \centering
    \caption{Example high-gap instances of a HAC model tested on TSPLib50. We see that all of these failure cases have large distances between node clusters, and thus deviate far from uniform instances.}
    \label{fig:prelim_hac_failures}
\end{figure*}

\subsection{TSPLib50 Dataset}

We motivate the creation of a new testing dataset, which we call TSPLib50. An issue with datasets generated from uniform distributions or other synthetic distributions is that they are not realistic, and thus not necessarily representative of real-world use cases of TSP solvers. A solution would be to test on TSPLib, a collection of real-world instances of the TSP, which is often used as a benchmark for combinatorial optimization solvers \cite{reinhelt_2014_tsplib}. Because TSPLib is based on real data, its distributions are both varied and relevant for real-world applications.

However, many solvers are trained on relatively small TSP instances. For example, \etal{Kool} \shortcite{kool_2018_attention} train and test on instances with 100 or fewer nodes. The models used by \etal{Zhang} \shortcite{zhang_2022_learning}, as well as our models in this paper, focus on TSP instances with 50 nodes. When tested on TSPLib, the gaps incurred by such models are not representative of generalization ability; instead, the gaps are correlated with instance size. This can be seen in Figure \ref{fig:prelim_hac_lengthgen}; there is a Pearson correlation of 0.907 between TSPLib instance size and resulting HAC model gap, demonstrating strong correlation. Improving model generalization to larger instance sizes often requires application of extensive computational resources, and is beyond the scope of preceding papers as well as this paper. However, this problem of instance size masks the issue of generalizing effectively to TSP instances from different \textit{distributions} within a particular instance size class.

Hence, we introduce TSPLib50, a dataset of 10,000 instances, each created by sampling 50 points uniformly at random from a TSPLib instance. Because the distribution of points in TSPLib50 is the same in expectation as that in TSPLib, we can thus disentangle generalization ability on different distributions with generalization ability on different instance sizes. Since TSPLib contains real-world instances, TSPLib50 evaluates robustness to realistic distributions.

\subsection{Hardness-Adaptive Curriculum Shortcomings}

While HAC as proposed by \etal{Zhang} \shortcite{zhang_2022_learning} constructs a clever curriculum and has thus made the most existing progress on improving performance for harder distributions, its coverage of ``hard" instances is far from exhaustive. Recall that a key component of HAC is the hardness-adaptive generator, where gradient ascent is conducted on training data. Because HAC samples training data from a uniform distribution and only conducts one step of gradient ascent on the sampled data, the distributions that the model ends up training on rarely deviate far from a uniform distribution.

We can see this in Equation \ref{eq:hardness_adaptive}: the amount each coordinate in $\bfX^{(t)}$ is modified is determined by its corresponding value in $\eta\nabla_{X^{(t)}} \calH(X^{(t)},M)$. From our experiments, we find that the mean value of elements inside $\eta |\nabla_{X^{(t)}} \calH(X^{(t)},M)|$ tends to be around $0.077$, while the median tends to be around $0.023$. These values are small relative to the unit square $[0,1]^2$ that points are placed in. Thus, training data is often only mildly perturbed by the hardness-adaptive generator, and indeed does not deviate far from the uniform distribution that they are sampled from.

We can confirm this by running experiments. We train a HAC model and test it on TSPLib50, and identify the instances with the highest gap. We plot the 4 instances with highest gap in Figure \ref{fig:prelim_hac_failures}, along with the oracle path and the path found by the HAC model. We can see that a main attribute these instances have in common is that they have a lot of empty space, corresponding to large distances between nodes; in other words, they are far from uniform distributions. Furthermore, we can see that the HAC model paths often make trivial mistakes such as crossing paths, which is indicative of lack of training on similar samples. This is supported by our previous analysis on how training samples cannot deviate far from uniform distributions because the training data is all sampled from uniform distributions.

This behavior is undesirable because TSPLib50 bootstraps from real-world distributions, and thus represents real-world use cases that we may care about. We seek to address this robustness issue in this paper.

\subsection{Other Evaluation Datasets}

While TSPLib50 evaluates performance on realistic distributions, we will also test performance of our method on synthetic distributions that are intended to be challenging. We test on the Gaussian Mixture distribution, which tends to pose a challenge to existing TSP solvers \cite{zhang_2022_learning}.

We also propose to test on a ``Diagonal" distribution of our design, which is intended to be difficult in another manner. Previously, we identified a common feature of HAC failure cases being that they have much empty space, and we justified this interpretation mathematically. However, another common feature of those cases is that they have points in distinct clusters. The Diagonal distribution aims to experimentally demonstrate that empty space is a primary factor for difficulty, by having all points aligned along a main diagonal. As such, there is only one cluster, but there is still much empty space on the TSP instance.

Visualizations of the TSPLib50, Gaussian Mixture, and Diagonal distributions can be seen in \ref{fig:distribution_visualization}. Technical details of their generation procedures are in the Appendix.

\begin{figure}[t]
    \includegraphics[width=0.5\textwidth]{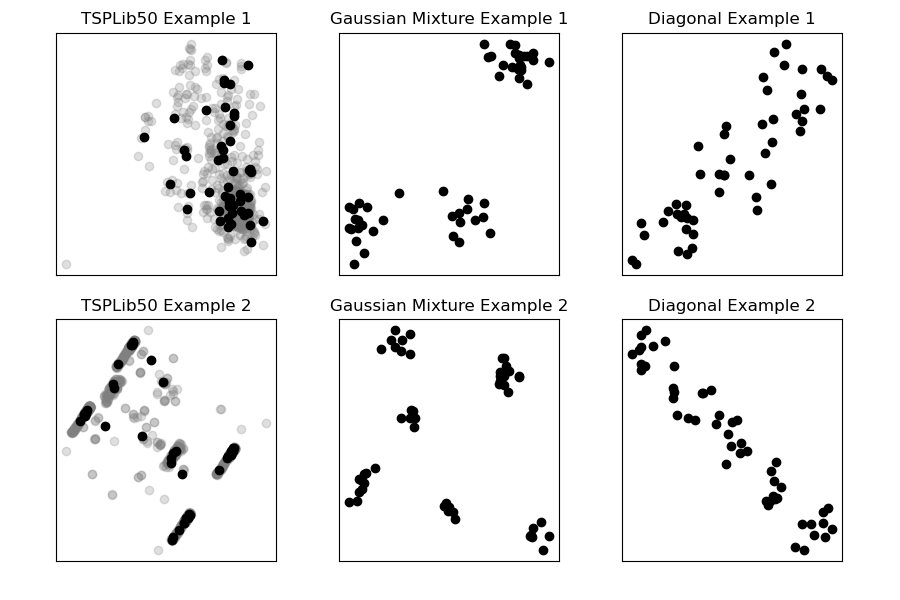}
    \centering
    \caption{Visualizations of example instances from the distributions we evaluate on. For TSPLib50, we also plot the original TSPLib instance that we sampled from in light gray.}
    \label{fig:distribution_visualization}
\end{figure}

\section{Method}

\begin{algorithm}[tb]
    \caption{CO with Generative Sampling (COGS)}
    \label{alg:gen_sampling}
    \textbf{Input}: Current model $M$, baseline model $M'$, VAE decoder $D$, validation dataset $\bfV$, hardness-adaptive generator $\phi$\\
    \textbf{Parameter}: Batch size $B$, training epochs $L$\\
    \textbf{Output}: Fine-tuned model $M'$
    \begin{algorithmic}[1] 
        \STATE Initialize and warm up $M$ and $M'$
        \FOR{$i=1,2,\ldots,L$}
            \STATE Sample vectors $z\sim\calN(0,1)$
            \STATE $\bfD \leftarrow D(z)$ (decode train dataset)
            \STATE $\bfD' \leftarrow \phi (\bfD')$ (pass dataset through HAC)
            \FOR{$b=1,2,\ldots,|\bfD|/B$}
                \STATE Get batch data $\{\bfX\}_{i=1}^B$ from $\bfD'$
                \STATE Pass batch data through baseline model $M'$
                \STATE Pass batch data through model $M$
                \STATE Update model parameters with weighted gradients
            \ENDFOR
            \IF{$\calC_M(\bfV) < \calC_{M'}(\bfV)$}
                \STATE $M' \leftarrow M$
            \ENDIF
        \ENDFOR
        \STATE \textbf{return} $M$
    \end{algorithmic}
\end{algorithm}

\subsection{Generative Model}

Since suboptimal performance on difficult distributions can be attributed to lack of coverage in training data, the intuitive solution is to train deep RL solvers on more diverse data. It is possible to hand craft diverse distributions to train on, but such distributions would be limited to the specific modes that are accounted for. However, previous work has shown that using a generative model to create training data for an RL agent can result in better coverage of a distribution than using the same training data to directly train the RL agent \cite{liang_2024_gamma}.

We choose to use a variational autoencoder (VAE) as our generative TSP model. VAEs are stochastic variational inference models which learn a distribution over latent representations \cite{kingma_2013_vae}. We use a VAE which encodes inputs into the parameters of a multidimensional Gaussian distribution, from which samples can be effectively drawn and decoded. This well-structured latent space allows for continuous data encoding, which results in interpolation within and extrapolation beyond the training dataset. In other domains, VAEs have been effectively used to generate data that can interpolate between and thus diversify the inputs it was trained on \cite{liang_2024_gamma}.

We use a sequence-to-sequence VAE architecture, where TSP instances are treated as sequences of points. To facilitate training, we sort TSP sequences to be monotonically increasing by $x$-coordinate. By treating TSP instances as sequences, the VAE can learn to reason about the distance relationships between points, which is a general representation capable of capturing the characteristics of a wide range of distributions. The VAE uses a long short-term memory (LSTM) \cite{hochreiter_1997_lstm} architecture for both the encoder and decoder. A diagram of our generative model architecture can be seen in Figure \ref{fig:architecture}.

We train the VAE on samples from a ``clustered uniform distribution" of our own design, which is intended to address the shortcomings of prior methods as found in our preliminary study. Intuitively, the clustered uniform distribution generates points in uniform clusters of variable size. In the special case where all clusters are of size 1, the distribution becomes a uniform distribution. We choose to train the VAE on a hand designed distribution as opposed to real-world data, so that the real-world TSPLib50 can be reserved solely for testing. However, in real-world deployments, training the VAE directly on test distributions may provide better results. Technical details of the clustered uniform distribution, VAE training, and related hyperparameters are in the Appendix.

\subsection{Training Distribution Coverage}

\begin{figure}[t]
    \includegraphics[width=0.5\textwidth]{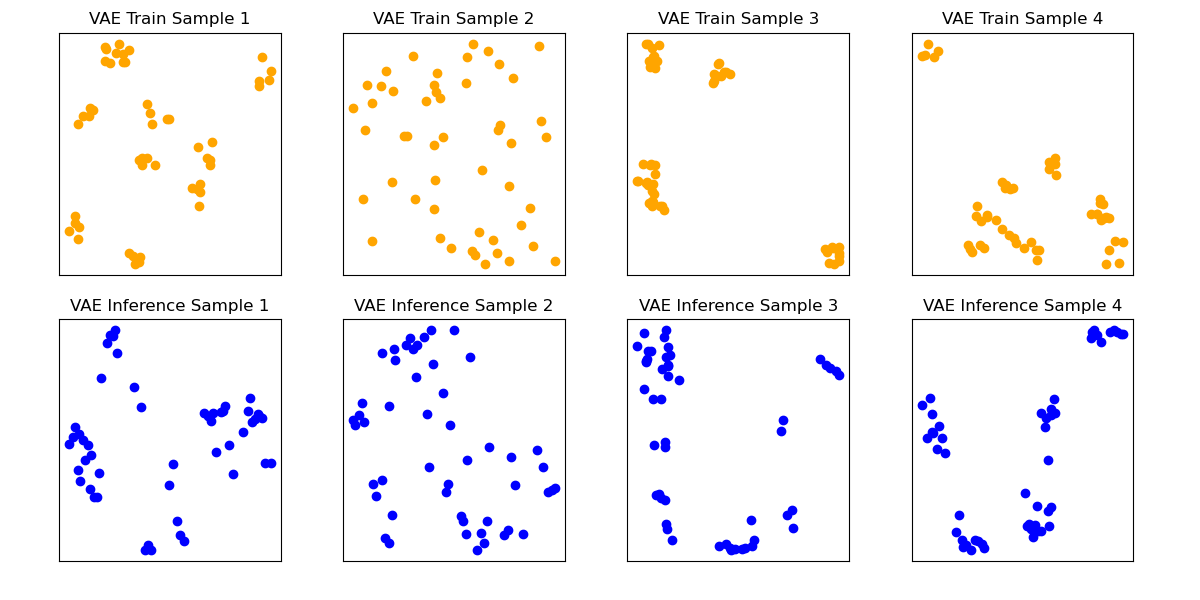}
    \centering
    \caption{Visualizations of VAE training samples (above) and VAE inference samples (below). VAE inference samples are able to ``interpolate" between training samples, thus expanding the training distribution and enabling the training of robust solvers.}
    \label{fig:vae_visualization}
\end{figure}

We argue that by using a VAE to generate training data for our TSP solver, we cover a greater range of possible TSP distributions than if we had trained on the clustered uniform distribution directly. Example instances of VAE training samples and VAE inference samples are visualized in Figure \ref{fig:vae_visualization}. Note that all VAE training samples are generated from the clustered uniform distribution. We observe that the VAE learns to interpolate between training instances and thus cover a wider range of possible distributions. For example, we see that each training instance in Figure \ref{fig:vae_visualization} is visibly either a clustered distribution or a uniform distribution. However, the VAE is able to generate samples that are a mix between a clustered and a uniform distribution, with some tight clusters and some regions that are more spread out. This can be seen most clearly in inference samples 1 and 2 from Figure \ref{fig:vae_visualization}.

\begin{figure}[t]
    \includegraphics[width=0.35\textwidth]{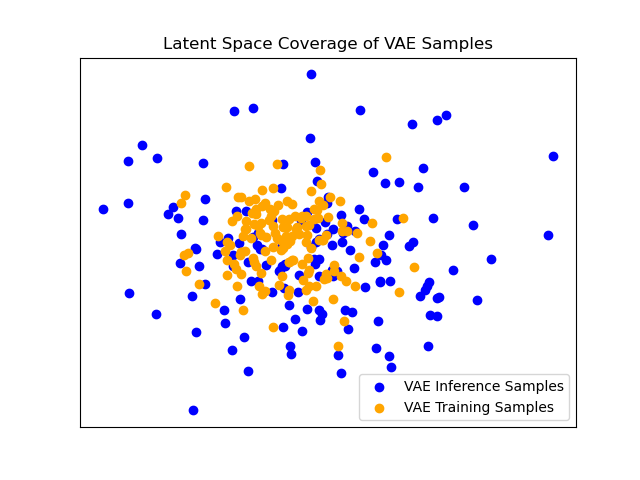}
    \centering
    \caption{PCA projection of the latent encoding of 150 VAE training samples and 150 VAE inference samples. VAE inference samples appear to cover of the latent space than training samples.}
    \label{fig:vae_coverage}
\end{figure}

This property is further supported by visualizing the latent space coverage of VAE training samples and VAE inference samples. Figure \ref{fig:vae_coverage} first encodes training and inference samples into latent space $z$, and then projects these latent representations into 2D using principal component analysis. It can be seen that VAE inference samples appear to cover more of the latent space than VAE training samples, which supports our hypothesis that using a VAE to generate training distributions allows better coverage and extrapolation in the space of possible TSP instances.

\subsection{Combinatorial Optimization with Generative Sampling (COGS)}

We propose Combinatorial Optimization with Generative Sampling (COGS) for robust optimization of deep RL solvers for the TSP, where we sample training data every epoch from our generative model. To design the architecture of the RL agent that will solve each TSP instance, we follow the work of \etal{Kool} \shortcite{kool_2018_attention}, we use an attention-based architecture trained with a REINFORCE rollout baseline. To create the RL objective, we follow the work of \etal{Zhang} \shortcite{zhang_2022_learning}, and use a hardness-adaptive curriculum and gradient re-weighting curriculum. We build off of these works to achieve state-of-the-art robustness on distributions of practical interest. Our key research question is whether improving the distribution of training data the RL agent receives can improve robustness.

During training, we first sample from our prior distribution latent vectors $z\sim \calN(0,1)$. Then, $z$ is passed as input to the decoder $D$ from our generative TSP model, which decodes $z$ into TSP training instances. The training instances are then passed through the hardness-adaptive generator $\phi$, which conducts a step of gradient ascent to make the instance harder. Afterwards, a tour is generated using an attention architecture and the loss with respect to network parameters $\theta$ is computed with REINFORCE:

\begin{equation}
    \calL_\theta (\bfX) = \doubleE_{p_{M_\theta}(\pi\mid\bfX)} \left[\calC (\pi\mid\bfX)\right]
    \label{eq:reinforce}
\end{equation}

The loss is then re-weighted, as is done by \etal{Zhang} \shortcite{zhang_2022_learning}, and model parameters are updated. COGS is specified in Algorithm \ref{alg:gen_sampling} and visualized in Figure \ref{fig:architecture}.

\section{Experiments}

\subsection{Baselines}

All our experiments work on fine-tuning a model previously trained exclusively on uniform random distributions. Notably, all experiments are run on a single GPU, and no model takes longer than a few hours to train. Experimental details and hyperparameters are in the Appendix.

We compare our model against two baselines: a uniform baseline that trains on uniform distributions without modification, and a HAC baseline that trains on uniform distributions passed through HAC. Note that these baselines are respectively equivalent to the work of \etal{Kool} \shortcite{kool_2018_attention} and the work of \etal{Zhang} \shortcite{zhang_2022_learning}. HAC is the current state-of-the-art on deep RL distribution generalization on the TSP.

In addition to the above, we also present ablations for two aspects of COGS: our use of a VAE and our use of HAC. The VAE ablation will train directly on the data used to train the VAE. The HAC ablation will not pass training data through a hardness-adaptive curriculum before training on it. The ablations primarily focus on practical robustness, and thus concern worst-case TSPLib50 gaps.

We plot gaps of all models relative to Concorde solutions, as Concorde is an optimal solver. Following the work of \etal{Zhang} \shortcite{zhang_2022_learning}, we also focus on 50-node instances.

\subsection{Evaluation}

We present results for average gaps on the Gaussian Mixture, Diagonal, and TSPLib50 distributions. We also present results on worst-case 1\%, 0.5\%, and 0.1\% of gaps on TSPLib50, to demonstrate the robustness of COGS to challenging out-of-distribution test cases. This robustness is important because TSPLib50 is a realistic dataset, and in real-world applications, performance guarantees are important. For example, a hypothetical routing program might require a guarantee that their routes are within 10\% gap of optimality in at least 99.5\% of cases, because 10\% gap accounts for thousands of miles in gas and driver time.

For all distributions, 10,000 instances are sampled for evaluation. We also train 5 models for each setting, and report averaged results between the 5 models for each setting.

\begin{figure}[t]
    \centering
    \includegraphics[width=0.5\textwidth]{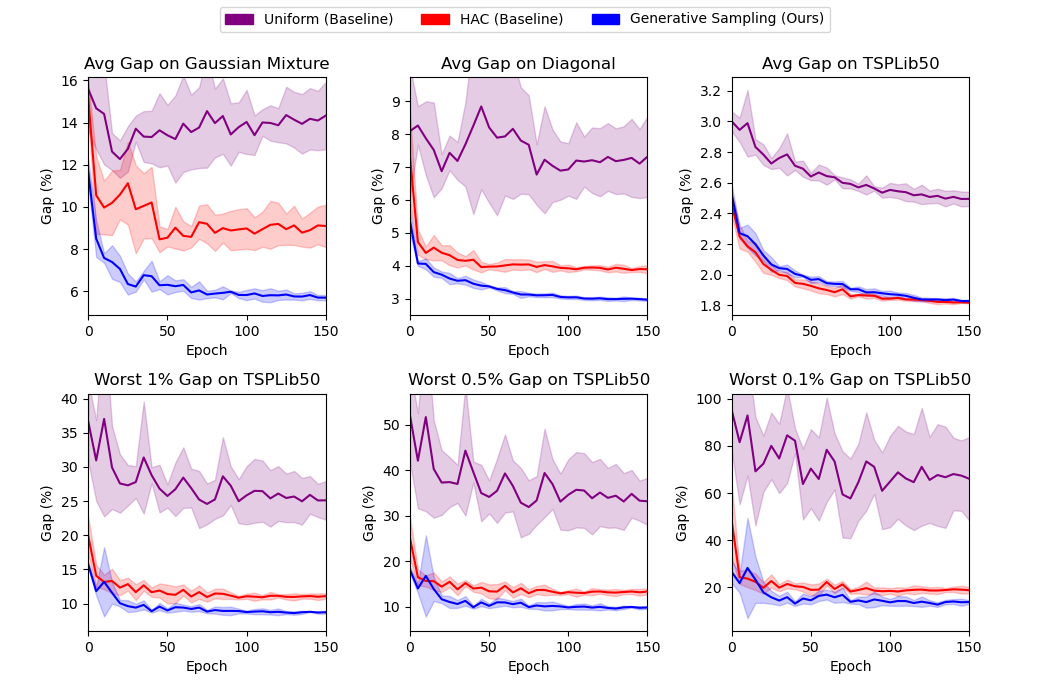}
    \caption{Gaps across training epochs of our proposed Generative Sampling model (blue) compared to baselines, on average cases across different distributions and worst-case scenarios in TSPLib50. We average optimality gaps across 5 random runs, with error bars representing standard deviation.}
    \label{fig:train_gap_main}
\end{figure}

\begin{table}[h!]
    \centering
    \begin{tabular}{l|l|l}
    \toprule
    Dataset & Model & Gap (\%) \\
    \midrule
    \multirow{3}{*}{\shortstack{Gaussian \\ Mixture}} & Uniform (Kool \shortcite{kool_2018_attention}) & 14.328 $\pm$ 1.591 \\
        & HAC (Zhang \shortcite{zhang_2022_learning}) & 9.099 $\pm$ 1.006 \\
        & \textbf{COGS (Ours)} & \textbf{5.704 $\pm$ 0.124} \\
    \midrule
    \multirow{3}{*}{Diagonal} & Uniform (Kool \shortcite{kool_2018_attention}) & 7.315 $\pm$ 1.223 \\
        & HAC (Zhang \shortcite{zhang_2022_learning}) & 3.897 $\pm$ 0.124 \\
        & \textbf{COGS (Ours)} & \textbf{2.973 $\pm$ 0.036} \\
    \midrule
    \multirow{3}{*}{TSPLib50} & Uniform (Kool \shortcite{kool_2018_attention}) & 2.495 $\pm$ 0.046 \\
        & \textbf{HAC (Zhang \shortcite{zhang_2022_learning})} & \textbf{1.818 $\pm$ 0.005} \\
        & COGS (Ours) & 1.828 $\pm$ 0.009 \\
    \midrule
    \multirow{3}{*}{\shortstack{TSPLib50 \\ Worst 1\%}} & Uniform (Kool \shortcite{kool_2018_attention}) & 66.055 $\pm$ 17.641 \\
        & HAC (Zhang \shortcite{zhang_2022_learning}) & 18.722 $\pm$ 1.389 \\
        & \textbf{COGS (Ours)} & \textbf{13.670 $\pm$ 0.818} \\
    \midrule
    \multirow{3}{*}{\shortstack{TSPLib50 \\ Worst 0.5\%}} & Uniform (Kool \shortcite{kool_2018_attention}) & 33.219 $\pm$ 5.150 \\
        & HAC (Zhang \shortcite{zhang_2022_learning}) & 13.396 $\pm$ 0.639 \\
        & \textbf{COGS (Ours)} & \textbf{9.896 $\pm$ 0.295} \\
    \midrule
    \multirow{3}{*}{\shortstack{TSPLib50 \\ Worst 0.1\%}} & Uniform (Kool \shortcite{kool_2018_attention}) & 66.055 $\pm$ 17.641 \\
        & HAC (Zhang \shortcite{zhang_2022_learning}) & 18.722 $\pm$ 1.389 \\
        & \textbf{COGS (Ours)} & \textbf{13.670 $\pm$ 0.818} \\
    \bottomrule
    \end{tabular}
    \caption{Average model gap across distributions, as well as worst case gaps on TSPLib50.}
    \label{table:performance}
\end{table}

\section{Results}

\subsection{Average Gaps}

Average gap results can be seen in Figure \ref{fig:train_gap_main} and Table \ref{table:performance}. As expected, gaps tend to be higher for the Gaussian Mixture and Diagonal distributions, as they are designed to be difficult. The gap for the Diagonal distribution is still fairly high despite there only being one cluster, which empirically confirms our assertion that empty space on the instance tends to cause for harder distributions. The average gaps on TSPLib50 are far lower because it is a more realistic dataset, and thus contains a lot of easier distributions that incur low gap.

We first observe that on all distributions, HAC already improves significantly on the uniform baseline, as HAC trains on uniform distributions that have been put through a hardness-adaptive generator. COGS, our proposed method, achieves consistent improvement over HAC on the harder distributions. We observe an approximately 1/3 factor decrease in gap on both hard distributions: the gap decreases from 9.10\% to 5.70\% on Gaussian Mixtures, and from 3.90\% to 2.97\% on the Diagonal distribution.

For the TSPLib50 distribution, COGS performs comparable to HAC in terms of average gap. This makes sense because a large portion of TSPLib50 instances are easy, while COGS focuses on robustness to challenging instances. The vast majority of instances in TSPLib50 are easy, and both HAC and COGS perform near optimally on those.

We find decreases in average gap between HAC and COGS to be statistically significant for the harder distributions, with values of $p < 0.003$ in two-sample t-tests for both the Gaussian Mixture and Diagonal distributions.

\subsection{Worst-Case Gaps}

Worst-case TSPLib50 gap results can be seen in Figure \ref{fig:train_gap_main} and Table \ref{table:performance}. COGS greatly improves robustness by optimizing performance in the worst-case scenarios. On TSPLib50, COGS provides a 2.36\% gap improvement on the worst 1\% of cases, a 3.5\% improvement on the worst 0.5\%, and a 5.05\% improvement on the worst 0.1\%. This is significant because in large-scale real-world applications that route millions of TSP problems every day, this is a considerable fraction. For example, if Amazon routes 40,000 loads every day within North America, 2.5\% of routes would still be 1,000 loads.

We find decreases in worst-case TSPLib50 gap between HAC and COGS to be statistically significant, with values of $p < 0.0001$ in two-sample t-tests for worst 1\% and worst 0.5\%, and $p < 0.001$ for worst 0.1\%.

COGS also improves gap from 220.26\% by HAC to 69.61\% on the worst 1\% of Gaussian Mixture cases, and from 26.59\% by HAC to 12.47\% on the worst 1\% of Diagonal cases. These several-fold decreases demonstrate COGS's impact in the most challenging scenarios.

\begin{figure}[t]
    \centering
    \includegraphics[width=0.5\textwidth]{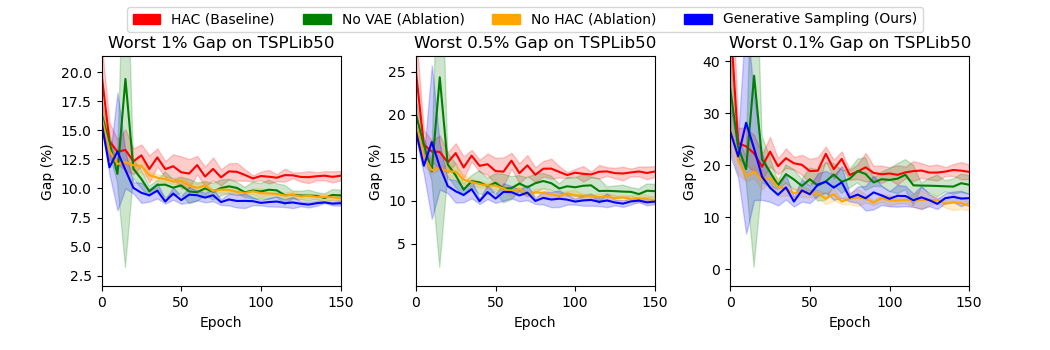}
    \caption{Gaps across training epochs of our proposed Generative Sampling model (blue) compared to ablations in practical robustness scenarios, as measured by worst-case scenarios in TSPLib50. HAC baseline is included for reference (red).}
    \label{fig:train_gap_ablation}
\end{figure}

\begin{table}[h!]
    \centering
    \begin{tabular}{l|l|l}
    \toprule
    \% Worst & Model & Gap (\%) \\
    \midrule
    \multirow{3}{*}{Worst 1\%} & \textbf{COGS (Ours)} & \textbf{8.733 $\pm$ 0.211} \\
        & No VAE & 9.404 $\pm$ 0.456 \\
        & No HAC & 9.155 $\pm$ 0.241 \\
    \midrule
    \multirow{3}{*}{Worst 0.5\%} & \textbf{COGS (Ours)} & \textbf{9.896 $\pm$ 0.295} \\
        & No VAE & 11.109 $\pm$ 0.796 \\
        & No HAC & 10.068 $\pm$ 0.356 \\
    \midrule
    \multirow{3}{*}{Worst 0.1\%} & COGS (Ours) & 13.670 $\pm$ 0.818 \\
        & No VAE & 16.250 $\pm$ 1.766 \\
        & \textbf{No HAC} & \textbf{12.192 $\pm$ 0.854} \\
    \bottomrule
    \end{tabular}
    \caption{Ablations for robustness on TSPLib50.}
    \label{table:ablation_avg}
\end{table}

\subsection{Ablations}

Ablation results for the VAE and HAC components of our method can be seen in Figure \ref{fig:train_gap_ablation} and Table \ref{table:ablation_avg}. It can be seen that if we do not use a VAE, and train our TSP solver directly on the same data used to train the VAE, there is a slight performance degradation in the worst case scenarios. This supports our hypothesis that the VAE learns to interpolate between training data, and thus generates a greater diversity of data than what it was trained on. Thus, the generative model is an integral component of our method with regard to robustness to the most challenging real-world cases.

If we do not use HAC, the performance drop is less significant, especially for worst 0.5\% and worst 0.1\%. This supports our preliminary study analysis on the importance of sampling distribution, and demonstrates how our generative model accounts for most of the performance improvement in COGS.

\section{Discussion}

\subsection{Conclusion}

We propose COGS, a training method for deep RL TSP solvers which samples training data from a generative model. By doing so, COGS covers a much broader range of distributions, allowing better generalization. We show that COGS significantly improves the performance of deep RL TSP solvers on hard distributions, especially in ``worst-case" scenarios on real-world distributions of practical interest. Such improved robustness and performance guarantees are significant in real-world deployment.

\subsection{Limitations}

We acknowledge that there exist realistic distributions that COGS still fails to consistently generate and train on, as noted by worst-case gaps that are still considerably suboptimal. Furthermore, to prevent test set contamination, we chose to hand craft a VAE training distribution as opposed to directly training on TSPLib50. In real-world deployments, performance may be further optimized if the VAE was trained directly on relevant real-world data.

This paper also focused on instances with 50 nodes. With more compute, COGS could be tested at scale on larger instance sizes. Additionally, while we conducted basic hyperparameter searches for our models and distributions, our search was not exhaustive. Thus, we believe that our performance could be optimized upon further tuning.

\subsection{Future Work}

COGS is a general methodology, and could be applied to other NCOs or COPs. As the \etal{Kool} \shortcite{kool_2018_attention} architecture generalizes to other problems such as the Vehicle Routing Problem (VRP) and Capacitated VRP (CVRP), it would be exciting to generalize COGS to other COPs.

\section{Acknowledgements}

This work was generously supported by an Amazon Middle Mile Product Tech Gift Research Award. We would like to thank Matthew Galati and Tim Jacobs for meaningful discussions that informed this work. We would like to thank Adrian Salamon for providing resources that helped us explore research in this area, including pointers to the open-source codebases that we built off of.

\bibliographystyle{named}
\bibliography{main}

\end{document}